\PassOptionsToPackage{unicode}{hyperref}
\PassOptionsToPackage{hyphens}{url}
\PassOptionsToPackage{dvipsnames,svgnames,x11names}{xcolor}
\documentclass[
  11pt,
]{article}
\usepackage{amsmath,amssymb}
\usepackage{iftex}
\ifPDFTeX
  \usepackage[T1]{fontenc}
  \usepackage[utf8]{inputenc}
  \usepackage{textcomp} 
\else 
  \usepackage{unicode-math} 
  \defaultfontfeatures{Scale=MatchLowercase}
  \defaultfontfeatures[\rmfamily]{Ligatures=TeX,Scale=1}
\fi
\usepackage{lmodern}
\ifPDFTeX\else
\fi
\IfFileExists{upquote.sty}{\usepackage{upquote}}{}
\IfFileExists{microtype.sty}{
  \usepackage[]{microtype}
  \UseMicrotypeSet[protrusion]{basicmath} 
}{}
\makeatletter
\@ifundefined{KOMAClassName}{
  \IfFileExists{parskip.sty}{%
    \usepackage{parskip}
  }{
    \setlength{\parindent}{0pt}
    \setlength{\parskip}{6pt plus 2pt minus 1pt}}
}{
  \KOMAoptions{parskip=half}}
\makeatother
\usepackage{xcolor}
\usepackage[margin=0.95in]{geometry}
\usepackage{longtable,booktabs,array}
\usepackage{calc} 
\usepackage{etoolbox}
\makeatletter
\patchcmd\longtable{\par}{\if@noskipsec\mbox{}\fi\par}{}{}
\makeatother
\IfFileExists{footnotehyper.sty}{\usepackage{footnotehyper}}{\usepackage{footnote}}
\makesavenoteenv{longtable}
\providecommand{\tightlist}{%
  \setlength{\itemsep}{0pt}\setlength{\parskip}{0pt}}
\usepackage{amsmath,amssymb}
\usepackage{booktabs}
\usepackage{array}
\usepackage{graphicx}
\usepackage{tikz}
\usetikzlibrary{shapes.geometric, arrows.meta, positioning}
\usepackage{newunicodechar}
\newunicodechar{⊕}{\ensuremath{\oplus}}
\newunicodechar{∧}{\ensuremath{\wedge}}
\newunicodechar{≪}{\ensuremath{\ll}}
\newunicodechar{≫}{\ensuremath{\gg}}
\newunicodechar{≈}{\ensuremath{\approx}}
\newunicodechar{→}{\ensuremath{\to}}
\newunicodechar{×}{\ensuremath{\times}}
\newunicodechar{≤}{\ensuremath{\le}}
\newunicodechar{≥}{\ensuremath{\ge}}
\newunicodechar{∞}{\ensuremath{\infty}}
\newunicodechar{∈}{\ensuremath{\in}}
\newunicodechar{⊥}{\ensuremath{\perp}}
\newunicodechar{⟺}{\ensuremath{\iff}}
\newunicodechar{⟹}{\ensuremath{\implies}}
\newunicodechar{⇒}{\ensuremath{\Rightarrow}}
\newunicodechar{↔}{\ensuremath{\leftrightarrow}}
\newunicodechar{…}{\ldots}
\newunicodechar{κ}{\ensuremath{\kappa}}
\newunicodechar{∝}{\ensuremath{\propto}}
\usepackage{etoolbox}
\AtBeginEnvironment{longtable}{\small}
\AtBeginEnvironment{tabular}{\small}
\AtBeginDocument{\let\OldUnderscore\_\renewcommand{\_}{\OldUnderscore\allowbreak}}
\ifLuaTeX
  \usepackage{selnolig}  
\fi
\IfFileExists{bookmark.sty}{\usepackage{bookmark}}{\usepackage{hyperref}}
\IfFileExists{xurl.sty}{\usepackage{xurl}}{} 
\hypersetup{
  pdftitle={Creative Integration: A Decidable Criterion of Creativity},
  pdfauthor={Yoshinori Nomura, Mirage Mountain Technologies Inc.~(nomura@miragemt.com)},
  colorlinks=true,
  linkcolor={Maroon},
  filecolor={Maroon},
  citecolor={Blue},
  urlcolor={Blue},
  pdfcreator={LaTeX via pandoc}}

\title{Creative Integration: A Decidable Criterion of Creativity}
\author{Yoshinori Nomura, Mirage Mountain Technologies
Inc.~(nomura@miragemt.com)}
\date{}

\begin{document}
\maketitle
\begin{abstract}
``Integrative'' solutions are widely praised but rarely defined: we lack
an operational way to tell a genuine integration --- one that makes the
world cheaper to describe --- from a tidy re-description. Building on
the lineage that treats creativity and intelligence as compression, we
give such a criterion for \emph{creative integration} (CI): the
resolution of a real conflict \(A \oplus B\) is CI if and only if, under
a fixed description language, the description length strictly shrinks
(\(C = L_{\text{pre}}/L_{\text{post}} > 1\)), with the reduction located
in the conflict itself. We make the judgment decidable through four
binary, conjunctive gates, and we fix its extension through a taxonomy
of pseudo-integration that names and rejects the look-alikes. We back
the criterion with a curated, multi-domain corpus and --- crucially ---
validate it not by human inter-rater agreement but by four falsifiable
tests it could fail: an independent computational check, discrimination
against hard negatives, and out-of-sample prediction --- which pass with
margin --- and description-language robustness, which holds for strong
compression and is reported honestly as bounded at the \(C\approx1\)
boundary. The contribution is not ``creativity is compression'' but its
decidability, discrimination, and corpus: on this account, what makes a
move \emph{genuinely} creative --- rather than merely novel --- is that
it compresses a conflict, with novelty and value as downstream symptoms;
whether \emph{all} creativity is so constituted we state as an explicit
conjecture. We claim only the sign of \(C-1\); we judge, not generate.
The result is a citable primitive for a broader program.
\end{abstract}

\hypertarget{introduction}{%
\section{§1 --- Introduction}\label{introduction}}

``Integrative'' solutions are praised across science, design, and
engineering: a theory that unifies two phenomena, an architecture that
reconciles two competing demands, a move that dissolves an apparent
trade-off. Yet praise outruns criteria. We have no operational way to
tell a \emph{genuine} integration --- one that makes the world cheaper
to describe --- from a showy re-description that merely narrates two
things as one. Without such a criterion, ``this is an elegant
integration'' is an aesthetic judgment, not a claim that can be checked.

\textbf{The gap.} There is a well-developed lineage holding that
creativity and intelligence are forms of compression (Schmidhuber's
formal theory of creativity; Hutter's compression-equals-intelligence;
the MDL/Kolmogorov foundations). That lineage gives the \emph{currency}
--- shorter descriptions --- but, for the question above, it stops short
in three ways: it offers (i) no operational, per-case criterion that
decides whether a given resolution is a true integration, (ii) no
discriminative boundary that separates real integrations from the many
look-alikes, and (iii) no labeled data on which either could be tested.
The thesis ``creativity is compression'' is, by itself, not yet a usable
judgment.

\textbf{This paper.} We close that gap for the specific phenomenon of
\emph{creative integration} (CI): the resolution of a genuine conflict
\(A \oplus B\) by a unifying principle under which the description
length strictly shrinks. Our contributions are:

\begin{enumerate}
\def\labelenumi{\arabic{enumi}.}
\tightlist
\item
  \textbf{An operational definition.} CI holds iff, under a declared
  description language, the compression ratio
  \(C = L_{\text{pre}}/L_{\text{post}} > 1\), estimated by a
  four-category counting procedure (principles / parameters / exceptions
  / boundaries) whose reduction is \emph{located in the conflict itself}
  --- the boundary and exception terms vanish (§3).
\item
  \textbf{A binary judgment rubric.} Four ordered, conjunctive gates
  (conflict-real, not-orthogonal, compression, locus-is-conflict); the
  first gate that fails decides the verdict and names the failure mode
  (§4).
\item
  \textbf{A discriminative boundary.} A taxonomy of
  \emph{pseudo-integration} --- cause-elimination, orthogonal axes,
  sequencing, enumeration, codification, calibration, standardization,
  packaging --- each anchored to the gate it fails, so the criterion is
  defined as much by what it rejects as by what it accepts (§5).
\item
  \textbf{A labeled corpus and measurement validity.} A curated,
  multi-domain corpus of 201 labeled cases (§7), validated not by human
  inter-rater agreement but by \textbf{measurement} --- four falsifiable
  tests the criterion could fail, on two axes: that the procedure
  measures \(C\) (computational check, language robustness) and that the
  verdicts are right (discrimination, out-of-sample prediction) (§8).
\end{enumerate}

\textbf{What this says about creativity.} On this criterion, what
separates a \emph{genuinely} creative move from a merely novel one is
its integrative core --- the compression of a conflict; novelty and
value, the usual currency by which creativity is evaluated, are
downstream symptoms (§2). We advance this as the paper's reading of
creativity. Its strongest form --- that \emph{all} genuine creativity is
creative integration --- we state and examine as a falsifiable
conjecture, not a result, in §10.2.

\textbf{What we do not claim.} The novelty is not ``creativity is
compression'' --- that is the prior lineage. It is the
\emph{decidability}, the \emph{discrimination}, and the \emph{corpus}
that turn the thesis into a checkable criterion. We claim only the
\textbf{sign} of \(C-1\), not its magnitude; we judge, we do not
generate (a generation method is deferred to follow-on work); and we do
not here pursue the larger unification the program points toward ---
that creativity, life, and intelligence may share this structure. This
paper is a criterion and its validation --- a citable primitive on which
the rest of the program can build.

\hypertarget{related-work-lineage}{%
\section{§2 --- Related Work / Lineage}\label{related-work-lineage}}

Our criterion descends from a recognized line of work that treats
compression as the currency of creativity and intelligence; positioning
it there makes clear that it is neither an isolated idea nor a
restatement of that line, but a specific addition on top of it.

\textbf{Compression as creativity and intelligence.} Schmidhuber's
formal theory of creativity casts interestingness and creative behavior
as \emph{compression progress} --- the rate at which an observer's model
of the world shortens (Schmidhuber 2010); it is our nearest neighbor and
we share its vocabulary. Hutter's program makes the identification
explicit at the level of intelligence --- better compression is better
prediction is more intelligence --- and operationalizes it in the Hutter
Prize (Hutter 2005); this is the same currency as our \(C>1\). The
measurement of ``description length'' rests on
minimum-description-length modeling (Rissanen 1978) and on Kolmogorov
complexity (Kolmogorov 1965), which also supply the two caveats we must
answer: description length is language-relative and uncomputable in
general. We address both by fixing the description language per record
and claiming only sign-invariance, with a tractable pre/post counting
approximation rather than true \(K(\cdot)\) (§3, §8, §9).

\textbf{Combining frames: bisociation and conceptual blending.} The act
we formalize --- resolving two conflicting frames into one --- has been
described richly, but never decidably. Koestler's \emph{bisociation}
casts creativity as connecting two self-consistent yet habitually
incompatible ``matrices of thought'' (Koestler 1964); Fauconnier and
Turner's \emph{conceptual blending} (which they also call conceptual
integration) projects two mental spaces into a blend with emergent
structure (Fauconnier and Turner 2002) --- and even names
``compression'' as a governing principle, though theirs is the semantic
compression of \emph{vital relations} across spaces, not description
length. Both name our phenomenon, \(A\oplus B\), but describe \emph{how}
integration happens rather than deciding \emph{when} it is creative:
they separate neither a genuine integration from a routine blend nor
supply a measurable, decidable handle. We take their phenomenon and give
exactly that.

\textbf{What creativity is, not just how it looks.} Computational
creativity has largely operationalized creativity through its
\emph{symptoms} --- novelty and value --- and built evaluation on them:
Boden's combinational / exploratory / \emph{transformational} taxonomy
(Boden 2004), Wiggins's formalization of creative systems as
rule-governed search over a conceptual space (Wiggins 2006), and
criteria that score generative outputs for novelty and value (Ritchie
2007; Colton 2008; Jordanous 2012). But novelty and value are downstream
signs, not the thing itself --- noise is novel, and value is adjudicated
by the very human agreement that cannot settle what is genuinely
creative (§8). We propose instead a \emph{constitutive} reading: what
makes a move genuinely creative, rather than merely novel, is that it
\emph{compresses} --- it resolves a real conflict into a description
strictly shorter than holding its terms apart --- and creative
integration is that property made decidable. This inverts the usual
relation to the evaluation literature: we are not measuring a corner of
creativity but proposing a definition of its core, of which novelty and
value are consequences. Because the test is structural, it is also
agent-neutral (§3.2): it answers, decidably, a question that
perception-based evaluation cannot settle --- whether a \emph{machine's}
act is creative (Colton 2008) --- by judging the act's structure, not
the agent's intent (developed in §10.2). Boden's \emph{transformational}
creativity --- reshaping the rules of a space --- is the nearest prior
form; \(C>1\) says when such reshaping is creative rather than merely
different. What this paper establishes is the criterion and its
measurement validity for integration; whether \emph{all} genuine
creativity is so constituted we take up, as a falsifiable conjecture, in
§10.2.

\textbf{Explanatory unification.} In philosophy of science, the value of
a unifying theory has been analyzed as a kind of compression: as a
reduction in the number of independently accepted phenomena (Friedman
1974), and as the derivation of the most conclusions from the fewest,
most stringent argument patterns (Kitcher 1981, 1989). This is
conceptually adjacent to CI restricted to scientific theories, and a
natural bridge; we keep its full development for a later,
history-of-science paper and use only consensus scientific integrations
(e.g.~Maxwell) here as \emph{calibration} exemplars (§7), not as
historiographical claims.

\textbf{The delta, at a glance.} The prior frames describe \emph{how}
two frames combine; none decides \emph{when} the combination is a
genuine integration rather than a tidy re-description. The contrast is
sharpest in a table.

\begin{longtable}[]{@{}
  >{\raggedright\arraybackslash}p{(\columnwidth - 8\tabcolsep) * \real{0.2000}}
  >{\raggedright\arraybackslash}p{(\columnwidth - 8\tabcolsep) * \real{0.2000}}
  >{\raggedright\arraybackslash}p{(\columnwidth - 8\tabcolsep) * \real{0.2000}}
  >{\raggedright\arraybackslash}p{(\columnwidth - 8\tabcolsep) * \real{0.2000}}
  >{\raggedright\arraybackslash}p{(\columnwidth - 8\tabcolsep) * \real{0.2000}}@{}}
\toprule\noalign{}
\begin{minipage}[b]{\linewidth}\raggedright
\end{minipage} & \begin{minipage}[b]{\linewidth}\raggedright
Bisociation (Koestler 1964)
\end{minipage} & \begin{minipage}[b]{\linewidth}\raggedright
Conceptual blending (Fauconnier \& Turner 2002)
\end{minipage} & \begin{minipage}[b]{\linewidth}\raggedright
Transformational creativity (Boden 2004)
\end{minipage} & \begin{minipage}[b]{\linewidth}\raggedright
\textbf{Creative integration (this work)}
\end{minipage} \\
\midrule\noalign{}
\endhead
\bottomrule\noalign{}
\endlastfoot
Core move & connect two habitually incompatible ``matrices of thought''
& project two mental spaces into an emergent blend & reshape the rules
of a conceptual space & resolve a real \(A\oplus B\) so the world is
cheaper to describe \textbf{at the locus} \\
Describes \emph{how} or decides \emph{when} & how & how & how &
\textbf{decides \emph{when}} (a verdict, not a mechanism) \\
Decidable verdict & no & no & no & \textbf{yes --- four binary,
conjunctive gates (§4)} \\
Names \& rejects look-alikes & no & no --- \emph{any} blend qualifies &
no & \textbf{yes --- nine-pitfall taxonomy (§5)} \\
Measurable handle & none & ``compression'' of \emph{vital relations}
(semantic, not \(L\)) & none & \textbf{sign of \(C-1\) under a fixed
description language (§3)} \\
Agent-neutral & implicit & no & no & \textbf{yes --- judges the act, not
the agent (§3.2, §10.2)} \\
Labeled corpus to test against & no & no & no & \textbf{yes (§7)} \\
\end{longtable}

The point is not that the prior frames are wrong --- they correctly
identify the phenomenon \(A\oplus B\) --- but that they are
\emph{generative and permissive} where ours is \emph{discriminative and
decidable}. Their strength is describing the mechanism of combination;
ours is saying, of a given case, \textbf{no} --- separating a genuine
integration from a routine blend, a re-labeling, or a packaged set of
decisions. Conceptual blending even invokes ``compression,'' but of
semantic vital relations across spaces, not of description length, and
with no boundary that a non-integration could fail.

\textbf{The delta.} None of these lines provides what a practitioner
needs to \emph{judge a given case}: a per-case operational criterion, a
discriminative boundary that names and rejects the look-alikes, or a
labeled corpus against which the criterion can be tested. The
compression thesis tells us what to value; it does not, on its own,
decide whether a particular resolution is a true integration or a tidy
re-description. Supplying that decidable criterion, its discriminative
taxonomy, and the data to validate it --- by measurement rather than by
agreement --- is the contribution of this paper.

\hypertarget{definition-ci-as-compression}{%
\section{§3 --- Definition: CI as
Compression}\label{definition-ci-as-compression}}

We define Creative Integration as an event in description length:
integrating two conflicting forces is \emph{creative} when the result
explains at least as much while costing strictly less to describe. This
section makes that precise, gives the counting handle we use to measure
it, and fixes the one degree of freedom an information-theoretic reader
will ask about --- the description language.

\hypertarget{description-length-decomposition}{%
\subsection{3.1 Description-length
decomposition}\label{description-length-decomposition}}

Let \(A\) and \(B\) be two forces that resist simultaneous maximization
under some pre-integration framing, written \(A \oplus B\). Following
minimum-description-length modeling (Rissanen 1978; Kolmogorov 1965), we
measure a system by its description length \(L(\cdot)\) under a fixed
language. Holding the unintegrated pair costs more than \(L(A)+L(B)\):
keeping two accounts from colliding incurs bookkeeping, so

\[L(A \oplus B) = L(A) + L(B) + L(\text{boundary}) + L(\text{exceptions}),\]

where \(L(\text{boundary})\) pays for the rules saying \emph{which
account applies when} and \(L(\text{exceptions})\) pays for itemizing
phenomena neither covers cleanly. These two terms are the signature of
an unintegrated conflict: each new configuration risks a new boundary
rule or exception.

Two notes on \(\oplus\). It is \textbf{notation for this conflicted
pair, not an algebraic operation}: it is meaningful only on forces that
genuinely compete, so it carries no identity and no idempotence (a
non-competing pair has \(L(\text{boundary})=L(\text{exceptions})=0\) and
is caught at gate G2, §4). And it is \textbf{super-additive} ---
\(L(A\oplus B) > L(A)+L(B)\) --- \emph{precisely because} they compete;
that excess is the conflict cost. The operational content of CI lies not
in \(\oplus\) but in re-describing \(A\oplus B\) as a single account
\(C\) (§3.2), a transition measured by \(L\) under a faithful
interpretation (§3.4); \(\oplus\) merely names the pre-integration
state.

\hypertarget{the-ci-condition}{%
\subsection{3.2 The CI condition}\label{the-ci-condition}}

A creative integration introduces a unifying principle under which the
boundary and exception terms vanish --- the two forces turn out to be
one thing seen from two sides. Writing
\(L_{\text{pre}} = L(A \oplus B)\) and \(L_{\text{post}}\) for the
unified account \emph{in the same language}, we summarize the reduction
by the compression ratio

\[C = \frac{L_{\text{pre}}}{L_{\text{post}}}, \qquad \text{CI} \iff C > 1,\]

observably so, with the reduction located in
\(L(\text{boundary})+L(\text{exceptions})\).

Two qualifiers carry weight. CI is \textbf{binary}, not a graded quality
score: a candidate either crosses the four gates of §4 (of which \(C>1\)
is G3) or it does not, and \(C \approx 1\) is a tidier restatement, not
CI. And the \textbf{locus} matters: a reduction obtained by bundling
decisions under one policy, while boundaries and exceptions re-appear
case by case, is packaging, not integration --- this is what gate G4
checks (§4, §5).

The definition is \emph{neutral to the integrating agent}. Whether the
conflict is compressed by a human reasoner, by a cultural process, or by
nature itself, only the structure --- a real conflict, compressed at its
locus --- decides. We tag this agent as the \emph{actor} ∈ \{human,
natural, cultural\}: it records \emph{who} integrated, never
\emph{whether} the result counts. Hence ``it is natural, therefore not
CI'' is not available as a verdict (§4.2), and the corpus spans all
three actors (§7), with the natural and cultural cases developed in
follow-on work.

\hypertarget{four-category-counting-the-operational-handle}{%
\subsection{3.3 Four-category counting (the operational
handle)}\label{four-category-counting-the-operational-handle}}

\(L(\cdot)\) is Kolmogorov-uncomputable in general, but the comparison
we need --- pre vs.~post in one language --- is tractable through a
\emph{counting approximation}: we tally four categories before and
after, each as a fixed count \textbf{or a scaling order} in a
problem-size parameter \(N\).

\begin{longtable}[]{@{}
  >{\raggedright\arraybackslash}p{(\columnwidth - 2\tabcolsep) * \real{0.5000}}
  >{\raggedright\arraybackslash}p{(\columnwidth - 2\tabcolsep) * \real{0.5000}}@{}}
\toprule\noalign{}
\begin{minipage}[b]{\linewidth}\raggedright
Category
\end{minipage} & \begin{minipage}[b]{\linewidth}\raggedright
What is counted
\end{minipage} \\
\midrule\noalign{}
\endhead
\bottomrule\noalign{}
\endlastfoot
\textbf{principles} & axioms, basic laws, governing assumptions \\
\textbf{parameters} & free parameters, per-configuration specifications,
initial/boundary data \\
\textbf{exceptions} & ``except when \ldots{}'' phenomena requiring
individual treatment \\
\textbf{boundaries} & rules selecting which account/method applies
where \\
\end{longtable}

\textbf{Worked example --- Maxwell (electricity \(\oplus\) magnetism).}
\emph{The conflict:} electricity and magnetism were separate force-laws
carrying a standing asymmetry --- a moving magnet and a moving conductor
produce the same current by \emph{different} rules --- together with a
clash between Ampère's law and charge conservation; the two genuinely
competed. \emph{The integration \& count:} one electromagnetic field in
four coupled equations. Counting in classical vector-field language:

\begin{longtable}[]{@{}
  >{\raggedright\arraybackslash}p{(\columnwidth - 4\tabcolsep) * \real{0.3333}}
  >{\raggedright\arraybackslash}p{(\columnwidth - 4\tabcolsep) * \real{0.3333}}
  >{\raggedright\arraybackslash}p{(\columnwidth - 4\tabcolsep) * \real{0.3333}}@{}}
\toprule\noalign{}
\begin{minipage}[b]{\linewidth}\raggedright
Category
\end{minipage} & \begin{minipage}[b]{\linewidth}\raggedright
Pre-integration (separate E and B)
\end{minipage} & \begin{minipage}[b]{\linewidth}\raggedright
Post-integration (one EM field, 4 equations)
\end{minipage} \\
\midrule\noalign{}
\endhead
\bottomrule\noalign{}
\endlastfoot
principles & ≈ 5--6 independent empirical laws (Coulomb/Gauss,
no-monopole, Biot--Savart, Ampère, Faraday) & one electromagnetic field,
\textbf{4 coupled equations} \\
parameters & a solution per source geometry, \(O(N)\) & boundary/initial
conditions, \(O(1)\) \\
exceptions & moving-magnet/moving-conductor asymmetry; Ampère vs.~charge
conservation & none standing (displacement current removes them) \\
boundaries & ``electric or magnetic?'' regime split & none --- E and B
are frame-dependent parts of one field \\
\textbf{total} & \(\approx 6 + O(N) + (\text{growing}) + 1\) &
\(4 + O(1)\) \\
\end{longtable}

The reduction lands exactly in \(L(\text{boundary})\) (the ``electric or
magnetic?'' split) and \(L(\text{exceptions})\) (the induction and
continuity anomalies) --- the locus the gates of §4 check. \emph{The
tell:} a phenomenon \emph{outside} the original conflict --- light as an
electromagnetic wave --- falls out as a prediction; genuine compression
buys new reach, a re-description does not.

\textbf{Recording format.} Each corpus record fixes a
\texttt{description\_language}, tallies the four categories for pre and
post as fixed counts or scaling orders, and states the ratio together
with a \emph{vividness signature} --- the single compression point
(here, the displacement-current term that erases the Ampère/continuity
exception and predicts light). We count \emph{generative} description
length --- the cost to reconstruct the phenomena, not the volume of data
they contain --- so a structure that derives \(N\) facts from \(O(1)\)
rules compresses even as the cases it covers grow.

\hypertarget{fixing-the-description-language}{%
\subsection{3.4 Fixing the description
language}\label{fixing-the-description-language}}

\(C\) is defined only relative to a language, and its magnitude is
language-dependent --- the Kolmogorov objection (e.g.~relativity is
short in tensor language, long in Newtonian). We therefore (i) declare
\texttt{description\_language} per record and count pre/post in the same
language, and (ii) claim only the invariance of
\(\operatorname{sign}(C-1)\) across admissible languages, never the
value of \(C\). For Maxwell, any reasonable physical language counts the
unified field as shorter; the \emph{sign} is what is determined.
Robustness of this sign is measured in §8 (T4) and the objection
answered in §9.

This sign-invariance is \textbf{provable, not assumed} --- a proposition
that, under an explicit finite-presentation idealization, follows from
the faithfulness of admissible re-descriptions. Model the pre- and
post-states as finitely presented theories and an \emph{admissible}
re-description as a \textbf{faithful interpretation} --- one preserving
derivability both ways (bi-interpretation). With \(\|\cdot\|\) the
minimal number of independent generators (case-stipulations folded in),
a faithful interpretation preserves that count up to a bounded overhead
\(d\) (the interpretation's defining schema, not the surface verbosity),
so the two compression gaps differ by at most \(2d\). Hence whenever the
structural compression exceeds \(2d\), \(\operatorname{sign}(C-1)\) is
invariant across admissible re-descriptions --- and, because each gate
of §4 is a predicate on the entailment structure, so is the \emph{entire
verdict}. The only escape, \(|\Delta|\le 2d\), is the marginal band the
rubric already routes to \emph{review}: a CI that a re-description could
un-sign is, by that token, not robust but borderline. Faithfulness is
also what blocks the adversarial ``smuggling'' language --- making the
post-state cheap by building the integration into a primitive forces the
pre-state more expensive under the same interpretation, so the gap
survives. (Statement and proof in the supplementary note on
sign-invariance.)

\hypertarget{scaling-analysis}{%
\subsection{3.5 Scaling analysis}\label{scaling-analysis}}

The deepest compression shows up as a lower scaling \emph{order}, not a
smaller constant. Let \(N\) index distinct electromagnetic
configurations. The pre-integration account needs configuration-specific
solutions and accretes an exception per new time-varying case ---
\(O(N)\); the four field equations generate every configuration from
boundary conditions --- \(O(1)\):

\[C \;\sim\; \frac{O(N)}{O(1)} \;\longrightarrow\; \infty \quad (N \to \infty),\]

and \(\gg 1\) even at fixed, modest \(N\). This \(O(N)\to O(1)\)
compression is the recurring fingerprint of CI across the corpus (§7),
and the same scaling test exposes \emph{enumerative} pseudo-integration
in §5 --- a ``unification'' whose specification grows with the problem
is not compression in disguise.

\hypertarget{the-judgment-rubric-four-gates}{%
\section{§4 --- The Judgment Rubric: Four
Gates}\label{the-judgment-rubric-four-gates}}

The definition of §3 becomes operational as a short, ordered checklist.
We make judgment \textbf{binary and conjunctive} rather than a graded
score, because each way of failing CI is a \emph{different, nameable}
error that a single number would blur. A candidate must pass four binary
gates; failing any one is disqualifying, and the identity of the failing
gate is the diagnosis --- it selects the pseudo-integration class of §5.
(This is kept separate from the corpus's auxiliary confidence scores,
which measure record quality, not the verdict; two records can tie on
them while the gates cleanly separate one as CI and the other as not.)

The definition is a single inequality --- CI ⟺ \(C>1\) (§3.2) --- so why
four gates rather than one? Because
\(C = L_{\text{pre}}/L_{\text{post}}\) is read off the counting
approximation of §3.3, and a number that \emph{looks} like \(C>1\) can
be inflated in four independent ways. Each gate closes one of them,
guarding a different part of the ratio:

\begin{longtable}[]{@{}
  >{\raggedright\arraybackslash}p{(\columnwidth - 4\tabcolsep) * \real{0.3333}}
  >{\raggedright\arraybackslash}p{(\columnwidth - 4\tabcolsep) * \real{0.3333}}
  >{\raggedright\arraybackslash}p{(\columnwidth - 4\tabcolsep) * \real{0.3333}}@{}}
\toprule\noalign{}
\begin{minipage}[b]{\linewidth}\raggedright
Gate
\end{minipage} & \begin{minipage}[b]{\linewidth}\raggedright
What part of \(C=L_{\text{pre}}/L_{\text{post}}\) it protects
\end{minipage} & \begin{minipage}[b]{\linewidth}\raggedright
The fake ``\(C>1\)'' it blocks
\end{minipage} \\
\midrule\noalign{}
\endhead
\bottomrule\noalign{}
\endlastfoot
\textbf{G1 conflict\_real} & the \textbf{numerator is a real conflict
cost} & a missing prerequisite inflated \(L_{\text{pre}}\); the
``reduction'' is cause-elimination, not integration \\
\textbf{G2 not\_orthogonal} & the numerator \textbf{has something to
compress} (\(L(\text{boundary})+L(\text{exceptions})>0\)) &
\(A\perp B\), so \(L_{\text{pre}}=L(A)+L(B)\) with no boundary/exception
term --- nothing to compress \\
\textbf{G3 compression} & the \textbf{ratio itself, and the denominator}
& \(L_{\text{post}}\) was shrunk by enumeration / codification /
calibration / sequencing --- an accounting trick, not a discovered
axis \\
\textbf{G4 locus\_is\_conflict} & \textbf{where} the reduction lands
(the ``located in \(L(\text{boundary})+L(\text{exceptions})\)'' clause
of §3.2) & the count is \(>1\) but by bundling decisions; boundaries and
exceptions re-appear case by case \\
\end{longtable}

So the gates are not four criteria competing with \(C>1\) --- they are
the four ways to fake it, each ruled out. \textbf{G1 and G2 secure the
numerator} (there is a genuine conflict cost to compress); \textbf{G3
measures the ratio and guards the denominator} (the shrink is a
discovered axis, not a trick); \textbf{G4 fixes the locus} (the shrink
is in the boundary/exception terms). Passing all four means \(C>1\)
\emph{honestly}: a real conflict, compressed at its locus. The failures
are nested --- there is no denominator worth checking before the
numerator is a real conflict --- so order matters, and the first failing
gate is the diagnosis that names the §5 pitfall.

\hypertarget{the-four-gates}{%
\subsection{4.1 The four gates}\label{the-four-gates}}

Each gate returns \texttt{\{pass,\ evidence\}}. The order matters: the
first two test whether the \emph{conflict is real at all}, the last two
whether a real conflict was actually \emph{compressed}.

\begin{longtable}[]{@{}
  >{\raggedright\arraybackslash}p{(\columnwidth - 4\tabcolsep) * \real{0.3333}}
  >{\raggedright\arraybackslash}p{(\columnwidth - 4\tabcolsep) * \real{0.3333}}
  >{\raggedright\arraybackslash}p{(\columnwidth - 4\tabcolsep) * \real{0.3333}}@{}}
\toprule\noalign{}
\begin{minipage}[b]{\linewidth}\raggedright
Gate
\end{minipage} & \begin{minipage}[b]{\linewidth}\raggedright
Question
\end{minipage} & \begin{minipage}[b]{\linewidth}\raggedright
If it fails → pitfall (§5)
\end{minipage} \\
\midrule\noalign{}
\endhead
\bottomrule\noalign{}
\endlastfoot
\textbf{G1 conflict\_real} & Would supplying a missing prerequisite make
the conflict disappear --- or is there no conflict being \emph{resolved}
at all, only a structure observed or described? &
\emph{cause-elimination} / \emph{interpretive discovery} \\
\textbf{G2 not\_orthogonal} & Do \(A\) and \(B\) actually compete, or is
``just do both'' the answer? & \emph{orthogonal axes} \\
\textbf{G3 compression} & Is \(L_{\text{post}} \ll L_{\text{pre}}\)
(\(C>1\)) via a discovered axis --- not enumeration, codification,
calibration, or sequencing? & \emph{sequencing} / \emph{enumerative
protocol} / \emph{codification} / \emph{standardization} /
\emph{calibration} \\
\textbf{G4 locus\_is\_conflict} & Is the locus of compression the
conflict itself --- do \(L(\text{boundary})\) and
\(L(\text{exceptions})\) vanish --- not a packaging of decisions? &
\emph{organizational packaging} \\
\end{longtable}

G4 is the subtle one: a candidate can show an apparent reduction yet
have its boundaries and exceptions re-appear case by case --- bundling
decisions under one policy without dissolving the tension.

\hypertarget{verdict-assignment}{%
\subsection{4.2 Verdict assignment}\label{verdict-assignment}}

Evaluation \textbf{short-circuits}: the first failure stops the process
and fixes the verdict and pitfall.

\begin{itemize}
\tightlist
\item
  \textbf{\emph{CI}} --- all of G1--G4 pass, \emph{and} the
  actor-specific generativity requirement holds (for
  \emph{human}/\emph{cultural}, competing descriptions genuinely
  pre-exist; for \emph{natural}, competing physical regimes genuinely
  co-exist). The \emph{actor} ∈ \{human, natural, cultural\} is always
  emitted but does not affect the verdict.
\item
  \textbf{\emph{not-CI}} --- any gate fails; the pitfall is the §5 name
  of the \emph{first} failing gate.
\item
  \textbf{\emph{review}} --- the record lacks information to decide a
  gate. The rule is conservative: \textbf{when unsure, do not tip to
  \emph{CI}; tip to \emph{review}.}
\end{itemize}

Two clarifications matter for G3: \(L(\cdot)\) is a \emph{generative}
description length --- a structure deriving \(N\) facts from \(O(1)\)
rules compresses even as the cases grow --- and \textbf{successful
prediction of unseen cases is evidence for compression, not against it}
(a compact structure recovering unseen cases at near-zero marginal cost
is exhibiting the regularity it compressed).

\hypertarget{maxwell-through-the-gates}{%
\subsection{4.3 Maxwell through the
gates}\label{maxwell-through-the-gates}}

\emph{The conflict:} electricity \(\oplus\) magnetism --- separate
force-laws that genuinely contend (a changing \(E\) \emph{produces}
\(B\)), carrying the induction asymmetry and the Ampère/continuity clash
(§3). \emph{Through the gates:} \textbf{G1} passes --- no added
prerequisite removes the asymmetry, so this is not cause-elimination;
\textbf{G2} passes --- \(E\) and \(B\) genuinely compete, not ``just do
both''; \textbf{G3} passes --- the unified field gives
\(\approx 6+O(N) \to 4+O(1)\) (§3.3); \textbf{G4} passes --- the
electric/magnetic boundary and the induction/continuity exceptions
\emph{disappear} rather than being repackaged. \emph{The verdict \&
tell:} all four pass ⇒ a \emph{CI} verdict, \emph{human} actor
(calibration set, §7); and the light prediction (§3.3) is the empirical
tell that the verdict is earned, not talked into.

\hypertarget{figure-gate-flowchart}{%
\subsection{4.4 Figure --- gate flowchart}\label{figure-gate-flowchart}}

\begin{figure}[ht]
\centering
\begin{tikzpicture}[
  node distance = 7mm and 6mm,
  font = \footnotesize,
  dec/.style  = {diamond, aspect=2.4, draw, align=center, inner sep=1pt},
  term/.style = {rounded corners, draw, align=center, inner sep=3pt},
  nc/.style   = {draw, align=center, inner sep=3pt, fill=black!4},
  ar/.style   = {-Stealth}
]
\node[term]            (S)  {candidate $A\oplus B$};
\node[dec, below=of S] (G1) {G1 conflict\\real?};
\node[dec, below=of G1](G2) {G2 not\\orthogonal?};
\node[dec, below=of G2](G3) {G3 $C>1$?};
\node[dec, below=of G3](G4) {G4 locus is\\the conflict?};
\node[term, below=of G4, fill=black!8](CI){verdict: \texttt{ci}};
\node[nc, right=of G1] (N1) {\texttt{cause\_elimination /}\\\texttt{interpretive\_discovery}};
\node[nc, right=of G2] (N2) {\texttt{orthogonal\_axes}};
\node[nc, right=of G3] (N3) {\texttt{sequencing / enumerative /}\\\texttt{codification / \ldots}};
\node[nc, right=of G4] (N4) {\texttt{organizational\_packaging}};
\draw[ar] (S) -- (G1);
\draw[ar] (G1) -- node[left]{yes} (G2);
\draw[ar] (G2) -- node[left]{yes} (G3);
\draw[ar] (G3) -- node[left]{yes} (G4);
\draw[ar] (G4) -- node[left]{yes} (CI);
\draw[ar] (G1) -- node[above]{no} (N1);
\draw[ar] (G2) -- node[above]{no} (N2);
\draw[ar] (G3) -- node[above]{no} (N3);
\draw[ar] (G4) -- node[above]{no} (N4);
\end{tikzpicture}
\caption{The four-gate decision flow. Evaluation short-circuits at the first failing gate, which names the pitfall (\S5); insufficient information yields \texttt{review} (omitted for clarity).}
\label{fig:gates}
\end{figure}
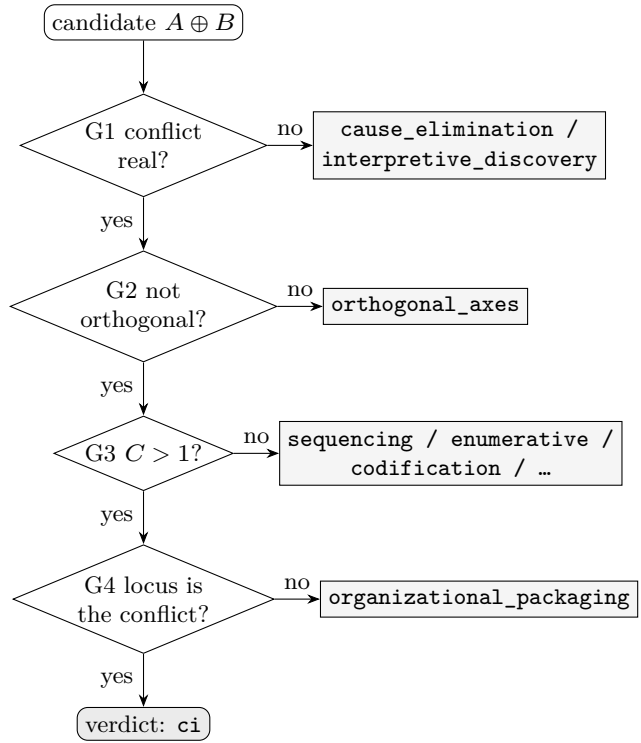

\hypertarget{the-discriminative-boundary-pseudo-integration}{%
\section{§5 --- The Discriminative Boundary:
Pseudo-Integration}\label{the-discriminative-boundary-pseudo-integration}}

A criterion that only confirms the obvious cases is a rubber stamp; the
discriminative power of CI lies in what it \emph{rejects}. The positive
condition (\(C>1\), located in the conflict) is easy to satisfy
\emph{rhetorically} --- almost any tidy re-description can be narrated
as ``unifying two things'' --- so the work is done by refusing the
narration when a gate fails. A rubric that fires \emph{CI} on
look-alikes measures fluency; one that rejects them discriminates. We
therefore define CI as much by its negative space --- a taxonomy of
pseudo-integration --- as by its positive condition, with each pitfall
anchored to the first gate it fails (§4). §8 (T2) reports the rejection
rates; here we fix \emph{what} must be rejected and \emph{why}.

\hypertarget{the-pitfall-taxonomy}{%
\subsection{5.1 The pitfall taxonomy}\label{the-pitfall-taxonomy}}

Representative classes (the table is illustrative, not exhaustive). Each
row follows the same three beats as the worked cases --- \emph{why it
looks integrative} (the symptom, with a real rejected example, largely
from AI governance and embodied skill), the \emph{gate it fails}, and
the \emph{tell}: what a genuine CI has that the look-alike lacks. Read
the symptom column if a particular example is unfamiliar.

\begin{longtable}[]{@{}
  >{\raggedright\arraybackslash}p{(\columnwidth - 6\tabcolsep) * \real{0.1800}}
  >{\raggedright\arraybackslash}p{(\columnwidth - 6\tabcolsep) * \real{0.4300}}
  >{\raggedright\arraybackslash}p{(\columnwidth - 6\tabcolsep) * \real{0.0900}}
  >{\raggedright\arraybackslash}p{(\columnwidth - 6\tabcolsep) * \real{0.3000}}@{}}
\toprule\noalign{}
\begin{minipage}[b]{\linewidth}\raggedright
Pitfall
\end{minipage} & \begin{minipage}[b]{\linewidth}\raggedright
Why it looks integrative (symptom --- example)
\end{minipage} & \begin{minipage}[b]{\linewidth}\raggedright
Fails
\end{minipage} & \begin{minipage}[b]{\linewidth}\raggedright
The tell: a CI instead\ldots{}
\end{minipage} \\
\midrule\noalign{}
\endhead
\bottomrule\noalign{}
\endlastfoot
\emph{cause-elimination} & the conflict vanishes once a missing
prerequisite is supplied --- RLHF: ``no reward signal'' dissolves once a
reward model is trained & G1 & \ldots has forces that still compete with
every prerequisite supplied \\
\emph{interpretive discovery} & a structure is merely observed, named,
or confirmed and narrated as an integration --- an observation
confirming an existing account, or a recurring pattern named across
cases; nothing is \emph{resolved}, so \(C\) is undefined & G1 &
\ldots{}\emph{resolves} a real \(A\oplus B\), not describes or confirms
a structure already there (integration event, not discovery) \\
\emph{orthogonal axes} & \(A\) and \(B\) co-exist, ``just do both''
works --- capability vs.~safety benchmarks (e.g.~HELM), no trade-off &
G2 & \ldots has \(A,B\) that cannot both be maximized --- a real
boundary term \\
\emph{sequencing} & an imposed time-order encoding nothing --- a drill's
phase labels (``recall / follow-through'') & G3 & \ldots lets the order
\emph{encode} an axis that compresses \\
\emph{enumerative protocol} & one abstract rule whose body is a per-case
spec growing with \(N\) --- a ``gate-above-threshold'' rule that is a
stack of per-tier specs & G3 & \ldots derives the cases from \(O(1)\)
rules, not enumerates them \\
\emph{codification} & an axis practitioners already used, written down
--- a risk×mitigation tiering policy & G3 & \ldots discovers a new axis,
not formalizes an old one \\
\emph{calibration} & parameters / mix-ratios tuned, no new control
variable --- fitting an inverted-U (arousal vs.~performance) & G3 &
\ldots introduces a new control variable \\
\emph{standardization} & an industry-wide format/spec agreed ---
coordination, \(L(A\oplus B)\) unchanged & G3 & \ldots reduces
\(L(A\oplus B)\), not just aligns conventions \\
\emph{organizational packaging} & decisions bundled under one policy ---
a release-tiering policy & G4 & \ldots dissolves the boundary; here it
re-appears per tier \\
\end{longtable}

(One further class is \emph{conditional} rather than a pure pitfall ---
applying a \emph{known} CI pattern to a new domain: it is admissible as
CI only if it re-achieves \(C>1\) in that domain and declares its
lineage; a bare re-application is codification + standardization.) By
contrast, Maxwell (§3--§4) survives every gate precisely \emph{because}
it is none of these: the boundary and exceptions are removed, not
codified, repackaged, or enumerated, and the unification predicts a new
phenomenon rather than re-labeling old ones.

\hypertarget{the-discrimination-test-rejecting-dissolved-paradoxes}{%
\subsection{5.2 The discrimination test: rejecting dissolved
paradoxes}\label{the-discrimination-test-rejecting-dissolved-paradoxes}}

A \emph{dissolved paradox} is an apparent conflict that, on inspection,
was never real --- the two sides do not actually compete, or the tension
disappears once a missing premise is supplied. The correct verdict is
\emph{not-CI} (nothing to integrate), and the rubric reaches it early,
at the G1/G2 gates that test whether the conflict is real at all. This
is a sharp test of the criterion itself: rejecting such a case shows it
is judging \emph{structure}, whereas accepting it shows it has been
fooled by \emph{presentation} --- rewarding a candidate for
\emph{sounding} paradoxical rather than for compressing a real conflict.
We therefore treat the rejection rate on dissolved paradoxes as a
first-class discrimination signal (§8, T2), alongside rejection of the
§5.1 pitfalls.

\hypertarget{operationalization-applying-the-gates-with-an-llm}{%
\section{§6 --- Operationalization: Applying the Gates with an
LLM}\label{operationalization-applying-the-gates-with-an-llm}}

The prior sections fix \emph{what} CI is and \emph{how} to test it ---
the four gates. This section is about how we actually run that test over
the corpus, and, just as importantly, what that procedure does and does
not establish.

Because the gates are binary, ordered, and mechanical, applying them is
a matter of following the rubric literally, not exercising taste. Each
record in the finished corpus (§7) carries a fixed, human-verified CI
label; the applicator never sees it. We embed the four-gate rubric of §4
in a prompt and run a language model \textbf{human-blind} --- given the
record's descriptive fields and the rubric alone, it must decide the
verdict for itself. For each record it emits a structured result ---
per-gate \texttt{\{pass,\ evidence\}}, the \texttt{verdict} (ci /
not\_ci / review), the \texttt{pitfall} on failure, the \texttt{actor}
(human / natural / cultural), and an advisory
\texttt{compression\_estimate} (description language, pre/post totals,
ratio) --- with a one-line rationale. This field list is the complete
output schema; reproducing the §8 numbers needs only the corpus (§7),
the frozen criterion text, and this schema. (That
\texttt{compression\_estimate} is advisory; the system of record for
\(C\) is the human four-category count of §3.3.)

We ran the rubric through several independent model families and the
verdicts were consistent. It is worth being exact about what that does
and does not buy, because two failure modes are easy to conflate:

\begin{itemize}
\tightlist
\item
  \textbf{Could the model simply tell us what we want?} This is what
  \emph{human-blind} application rules out: with the ground-truth label
  hidden, the applicator cannot parrot a desired answer --- it has only
  the fields and the rubric. Running many models would \emph{not} fix
  this on its own (models can share a leaning), so the control here is
  the blind, not the head-count.
\item
  \textbf{Does cross-model agreement prove the verdicts are right?}
  No.~A language model faithfully reproduces whatever rubric it is
  given, so agreement across models shows only that the rubric is
  \textbf{mechanically single-valued} --- an astrology rubric would
  agree with itself just as well. Agreement is therefore reported only
  as an auxiliary \emph{determinacy} signal (§8.5), never as validity.
\end{itemize}

Validity is carried instead by the four tests of §8, each pitting the
applicator against something it could genuinely disagree with. Two probe
\textbf{process validity} --- that the procedure measures \(C\) and not
an artifact of wording: an independent compression count (T1) and
paraphrased description languages (T4). Two probe \textbf{outcome
validity} --- that the verdicts are right and not curve-fit:
deliberately constructed hard negatives (T2) and unseen held-out cases
(T3). The LLM is the instrument that lets those tests run at scale ---
not the source of their authority.

The independence that test T1 needs --- a compression count separable
from the verdict --- is supplied by the four-category count of §3.3
(computed per record alongside, but not from, the gate verdict), not by
any separate machinery. (We explored a fully mechanical proxy ---
extracting each record to a semantic graph and reading \(C\) off it with
graph predicates --- but on a scaled corpus sample it did not reliably
reproduce the verdicts; that negative result is recorded in the
supplementary material. The validated operationalization is the
full-rubric, human-blind application of this section, checked against
the counts and held-out cases of §8.)

\hypertarget{the-corpus}{%
\section{§7 --- The Corpus}\label{the-corpus}}

The criterion is backed by a labeled corpus assembled in an internal
curation system. We describe its size, label distribution, domain
spread, and the calibration backbone that gives the rubric face
validity.

\hypertarget{scale-and-label-distribution}{%
\subsection{7.1 Scale and label
distribution}\label{scale-and-label-distribution}}

The corpus is the set of records that carry a finalized CI verdict:
\textbf{201 records --- 52 positives (\texttt{ci}) and 149 negatives
(\texttt{not\_ci})}. The positives are confirmed creative integrations;
the negatives are the boundary cases of §5. This is the dataset on which
every validity claim in §8 is computed --- the TPR and TNR denominators
of §8.2 are exactly these 52 and 149.

\hypertarget{the-negatives-carry-the-discrimination}{%
\subsection{7.2 The negatives carry the
discrimination}\label{the-negatives-carry-the-discrimination}}

A criterion defined by its boundary (§5) needs negatives, and the corpus
is negative-heavy by design: 149 of the 201 records are
\texttt{not\_ci}. Each negative is tagged with the pitfall class it
instantiates --- the human/cultural core being the §5 taxonomy:
cause\_elimination 98, codification 16, orthogonal\_axes 12, sequencing
9, enumerative\_protocol 7, organizational\_packaging 2, calibration 1,
plus the natural-actor look-alike \texttt{interpretive\_discovery} 4
(developed in follow-on work). These are the hard negatives that T2
(§8.2) must reject, and their spread across the taxonomy is what lets us
report \emph{per-pitfall} discrimination rather than a single aggregate.

\hypertarget{domain-spread-and-the-calibration-backbone}{%
\subsection{7.3 Domain spread and the calibration
backbone}\label{domain-spread-and-the-calibration-backbone}}

The 201 records span \textbf{61 domains}. The larger clusters include
western\_music (20), legal (14), piano\_skill (14), medical (12),
philosophy (12), organ\_evolution (12), painting (11), life\_evolution
(10), coaching (8), mathematics (7), and architecture (6). The breadth
supports the extension to nature's creations and the domain-generality
developed in follow-on work, but here it is deliberately \textbf{not}
the load-bearing claim (§8 reports a single clean validation; breadth is
for the broader program, §10).

Within the positives, a small \textbf{calibration backbone} does
specific work: consensus mathematical and chemical integrations
(mathematics 7, of which 6 \texttt{ci}; chemistry 1 \texttt{ci}) serve
as \emph{worked examples} on which the rubric must fire \texttt{ci}
without requiring domain-expert review --- they give face validity
cheaply. These are consensus cases any reader can check unaided; the
rubric fires \texttt{ci} on each and rejects the §5 look-alikes that
share their surface vocabulary of ``reconciling two forces'':

\begin{longtable}[]{@{}
  >{\raggedright\arraybackslash}p{(\columnwidth - 6\tabcolsep) * \real{0.2500}}
  >{\raggedright\arraybackslash}p{(\columnwidth - 6\tabcolsep) * \real{0.2500}}
  >{\raggedright\arraybackslash}p{(\columnwidth - 6\tabcolsep) * \real{0.2500}}
  >{\raggedright\arraybackslash}p{(\columnwidth - 6\tabcolsep) * \real{0.2500}}@{}}
\toprule\noalign{}
\begin{minipage}[b]{\linewidth}\raggedright
Consensus \texttt{ci}
\end{minipage} & \begin{minipage}[b]{\linewidth}\raggedright
Conflict \(A\oplus B\)
\end{minipage} & \begin{minipage}[b]{\linewidth}\raggedright
Unifying principle
\end{minipage} & \begin{minipage}[b]{\linewidth}\raggedright
\(C\)
\end{minipage} \\
\midrule\noalign{}
\endhead
\bottomrule\noalign{}
\endlastfoot
Maxwell (running example, §3--§5) & electricity ∧ magnetism & one EM
field, 4 equations & \(O(N)\to O(1)\) \\
Special relativity & space ∧ time & spacetime + invariant \(c\) & \(>3\)
(7 ether assumptions → 2 postulates) \\
Complex plane & algebra ∧ geometry & \(\mathbb{C}\) as the plane &
\(\gg 1\) \\
\end{longtable}

(Maxwell and special relativity serve as the paper's running examples
and as held-out worked examples in §8; the complex plane is among the
corpus's mathematics positives.)

\hypertarget{schema}{%
\subsection{7.4 Schema}\label{schema}}

Each record is stored across a relational schema (\textasciitilde15
tables) capturing the verdict/pitfall/actor fields, the axis annotations
(\(A\), \(B\), \(C\) and the construction), the construction mechanism,
the surface/abstract/root conflict levels, the integration, chronology,
and domain. The compression calculation follows the recording format of
§3.3.

\hypertarget{validation-measurement-validity}{%
\section{§8 --- Validation = Measurement
Validity}\label{validation-measurement-validity}}

\hypertarget{what-validity-means-here}{%
\subsection{8.0 What validity means
here}\label{what-validity-means-here}}

Humans err and language models err; a correct verdict is a probabilistic
outcome, not a guarantee. The question is therefore \textbf{not} whether
raters \emph{agree} --- agreement measures consensus, not correctness,
and human or model raters can agree on a wrong answer. (Cross-model
agreement is for this reason \emph{determinacy}, not validity; §6.) The
question is whether the rubric is a \textbf{valid measurement} of the
objective quantity it approximates, \(\operatorname{sign}(C-1)\).

A measurement is valid on two axes, and we test each directly:

\begin{itemize}
\tightlist
\item
  \textbf{Process validity} --- the procedure measures the intended
  quantity, not an artifact of how a case is phrased. \emph{Does the
  verdict track an independent computation of \(C\)} (\textbf{T1})?
  \emph{Is it invariant when the same structure is re-expressed}
  (\textbf{T4})?
\item
  \textbf{Outcome validity} --- the verdicts are correct, not curve-fit
  to the labels. \emph{Does the rubric reject hard look-alikes,
  including ones constructed after it was frozen} (\textbf{T2})?
  \emph{Does it hold on cases it never saw} (\textbf{T3})?
\end{itemize}

Each test is pre-registered: the criterion is frozen (commit hash), the
test set is fixed (hashed), and the passing bar is declared
\textbf{before} results are seen. Independence is enforced two ways ---
the applicator runs \textbf{human-blind} (§6), and the \emph{counting}
of \(C\) runs on a path \textbf{separate} from the \emph{gating} (the
counter never sees the verdict, the judge never sees the count).

Each test follows three beats --- what its \emph{failure} would mean
(the threat it guards), the pre-registered \emph{bar} it must clear, and
the \emph{measured} result (the tell, with its margin):

\begin{longtable}[]{@{}
  >{\raggedright\arraybackslash}p{(\columnwidth - 8\tabcolsep) * \real{0.2000}}
  >{\raggedright\arraybackslash}p{(\columnwidth - 8\tabcolsep) * \real{0.2000}}
  >{\raggedright\arraybackslash}p{(\columnwidth - 8\tabcolsep) * \real{0.2000}}
  >{\raggedright\arraybackslash}p{(\columnwidth - 8\tabcolsep) * \real{0.2000}}
  >{\raggedright\arraybackslash}p{(\columnwidth - 8\tabcolsep) * \real{0.2000}}@{}}
\toprule\noalign{}
\begin{minipage}[b]{\linewidth}\raggedright
Axis
\end{minipage} & \begin{minipage}[b]{\linewidth}\raggedright
Test
\end{minipage} & \begin{minipage}[b]{\linewidth}\raggedright
If it failed (the threat)
\end{minipage} & \begin{minipage}[b]{\linewidth}\raggedright
Pre-registered bar
\end{minipage} & \begin{minipage}[b]{\linewidth}\raggedright
Measured (the tell)
\end{minipage} \\
\midrule\noalign{}
\endhead
\bottomrule\noalign{}
\endlastfoot
\textbf{Process} & \textbf{T1} computational check & the verdict is a
vote, not a measurement of \(C\) & sign-agreement ≥ 0.85 & \textbf{1.00}
(primary), 0.933 (2nd family); inter-counter 0.933 \\
\textbf{Process} & \textbf{T4} robustness & the verdict tracks wording,
not structure & sign-invariance ≥ 0.90 & \textbf{0.85} {[}0.72, 0.93{]}
(\(n=50\)) --- stratified: \textbf{0.92} strong-CI / \textbf{0.78}
boundary \\
\textbf{Outcome} & \textbf{T2} discrimination & the rubric is a rubber
stamp & TNR ≥ 0.80, dissolved ≥ 0.95, TPR ≥ 0.80 & TNR \textbf{1.00},
dissolved \textbf{1.00}, TPR ≈ \textbf{1.00} \\
\textbf{Outcome} & \textbf{T3} prediction (OOS) & the rubric is a
post-hoc frame & held-out drop ≤ 10 pp & drop \textbf{≤ 1.6 pp} \\
\end{longtable}

T1--T3 pass with margin.\footnote{For \textbf{T2}, the \emph{Measured}
  column reports the expanded held-out set (hard negatives constructed
  after the criterion was frozen, plus held-out positives); the
  corpus-internal rates are lower --- TNR 0.986, TPR 0.865 --- and are
  given in §8.2.} \textbf{T4 is the one honest miss}: it passes for
strong compression (≈0.92) but falls below its 0.90 bar at the
\(C\approx1\) boundary (0.78), where invariance is bounded by framing
discipline rather than a measurement principle (§8.4, §9). The
subsections give each test's design and its honest caveat.

\hypertarget{t1-computational-check-process-validity}{%
\subsection{8.1 T1 --- Computational check (process
validity)}\label{t1-computational-check-process-validity}}

A blind counter fixes the \texttt{description\_language}, tallies the
four categories pre/post, and computes \(C\) \textbf{without seeing the
verdict}; the judge applies the gates \textbf{without seeing the count};
we compare \(\operatorname{sign}(C-1)\) against the verdict. Were the
verdict a vote rather than a measurement, the two paths would diverge
--- they did not. Sign-agreement was \textbf{15/15 = 1.00} (primary
family) and 14/15 = 0.933 (a second, independent family); inter-counter
agreement (a third party re-counting a sub-sample) was 0.933; and the
\(C\) distributions of \texttt{ci} and \texttt{not\_ci} separate cleanly
at \(1\). The single second-family disagreement is reported, not hidden;
the granularity threat is addressed by publishing the counting protocol
(§3.3) and the inter-counter rate.

\hypertarget{t2-discrimination-outcome-validity}{%
\subsection{8.2 T2 --- Discrimination (outcome
validity)}\label{t2-discrimination-outcome-validity}}

A rubber stamp scores TPR = 1, TNR = 0, so \textbf{TNR is the deciding
metric}. Hard negatives come from (a) the corpus's pitfall-tagged
\texttt{not\_ci} records and, crucially, (b) \textbf{newly constructed}
pseudo-integrations (codification, standardization, sequencing,
dissolved paradoxes) authored after the criterion was frozen --- so
success cannot be curve-fitting to the curator's labels; true CIs are
included for TPR. On the corpus, TNR = 0.986 / TPR = 0.865; on the
expanded held-out set, TPR = 21/21 ≈ \textbf{1.00}, TNR = 16/16 =
\textbf{1.00}, with \textbf{dissolved-paradox rejection = 1.00}. We
report corpus-internal and out-of-corpus rates separately so the
discrimination is visibly not an artifact of the labeling source.

\hypertarget{t3-prediction-out-of-sample-outcome-validity}{%
\subsection{8.3 T3 --- Prediction / out-of-sample (outcome
validity)}\label{t3-prediction-out-of-sample-outcome-validity}}

A post-hoc frame fits its own examples; a measurement holds on unseen
cases. On a 3-actor held-out set (human / natural / cultural, \(n=37\)),
held-out computational agreement was \textbf{37/37 = 1.00} (primary;
36/37 = 0.973 second) --- a \textbf{drop of ≤ 1.6 pp} against
development, well inside the ≤ 10 pp bar. Two clarifications: this is
the prediction of \emph{judgment}, not \emph{generation} (whether the
method can \emph{generate} integrations is a separate question, §10,
with a held-out of its own --- not to be conflated); and held-out
validity requires the discriminating evidence to be \emph{self-contained
in the visible fields} --- an earlier organ-evolution held-out failed
this (it measured input quality) and was re-curated (§8.5).

\hypertarget{t4-robustness-process-validity}{%
\subsection{8.4 T4 --- Robustness (process
validity)}\label{t4-robustness-process-validity}}

Against the Kolmogorov objection we claim only that
\(\operatorname{sign}(C-1)\) is invariant under an admissible class of
language changes --- natural-language swaps (JA↔EN), alternative valid
formalizations/granularities, re-expression by a different agent ---
forbidding information injection/deletion. We test this at scale
(pre-registered, frozen target hash): each of \(n=50\) balanced records
is re-expressed by a \emph{different} agent and \(C\) is re-counted on
\textbf{both} the original and the re-expression, so the counter's own
error cancels and only a sign change \emph{caused by} re-expression
counts. Invariance is \textbf{0.85} {[}0.72, 0.93{]}, and it is
\textbf{strongly stratified}: strong compression is robust (ci
\textbf{22/24 ≈ 0.92}; every flip sits at the margin --- none is a
\(C\gg1\) compression), while the \textbf{near-boundary
\texttt{not\_ci}} records, where \(|C-1|\approx0\), are labile
(\textbf{18/23 ≈ 0.78}). This \textbf{misses the pre-registered 0.90
bar}; we report it as a stratified result, not the small-\(n\) point
estimate it replaces.

The boundary flips are \textbf{definitional, not an irreducible
\(C\approx1\) principle}. When two faithful re-expressions of one
structure disagree, a gate term is under-specified --- most often ``the
boundary was \emph{eliminated}'' conflated with ``the frontier merely
\emph{moved}'' (after integration you can still trade \(A\) against
\(B\)). Operationalizing that single distinction, with its siblings
(codification, orthogonal axes, mere-description), and re-running raised
\texttt{not\_ci} invariance from \textbf{0.57 to 0.78}: defining the
term collapses the flip it governs. The gain is real but partial ---
sharper rejection tests also \emph{relocate} noise, tripping more
readily on near-boundary true CIs --- so this is a stack of definitional
fixes over an irreducible remainder, not a cure.

That remainder is \textbf{framing discipline}, which a sharper probe
isolates. Letting the applicator \textbf{re-choose} the frame from a
verdict-blind \texttt{\{name,\ situation\}} (\(n=60\), balanced, chance
0.5) gives the same asymmetry magnified: identification is frame-robust
(TPR \textbf{29/29 = 1.00}), rejection frame-fragile (TNR \textbf{0.70},
flips clustered on the pseudo-integration boundary). A disciplined
re-audit settles which side erred: holding each flipped record to its
\textbf{own existing description} (no favorable reframing) and requiring
\textbf{two independent judges to agree}, all 7 flips are applicator
\emph{permissiveness} --- a fabricated conflict where the record's own
structure is \texttt{complementary}/\texttt{commutative} --- not corpus
errors; 0/11 of the swept cluster certify. The discrimination discipline
(§4--§5) therefore lives in the \emph{choice of frame}, not the
phenomenon: the sign is robust for strong compression and
\textbf{bounded by framing discipline at the boundary} (§9). Tellingly,
the one contamination these audits surfaced was caught \textbf{by the
disciplined method itself}, not a human --- evidence the safeguard
binds.

\hypertarget{methodological-note-self-contained-held-outs}{%
\subsection{8.5 Methodological note: self-contained
held-outs}\label{methodological-note-self-contained-held-outs}}

A held-out tests the \emph{rubric} only if the fields visible to the
applicator contain the evidence needed to decide the gates; built on
fragile skeletons it silently tests input quality instead. The
organ-evolution held-out failed exactly this way; we surfaced it,
re-curated so the discriminating evidence is self-contained, and
recovered generalization --- documenting the negative result because it
is part of the validity argument. (Cross-model agreement ---
within-family 0.867, cross-family 15/15 = 1.00 --- is reported only as
determinacy, §6: it shows the rubric is single-valued, not that it is
right, and is subordinate to T1--T4. The residual circularity that
remains is addressed in §9.)

\hypertarget{threats-to-validity}{%
\section{§9 --- Threats to Validity}\label{threats-to-validity}}

Three objections come from information theory and one from research
methodology. Each follows three beats --- the \emph{objection},
\emph{why it misses}, and the \emph{tell}: the one claim under which it
would actually bite --- and is argued in full in the cited section
rather than restated here.

\begin{longtable}[]{@{}
  >{\raggedright\arraybackslash}p{(\columnwidth - 6\tabcolsep) * \real{0.2500}}
  >{\raggedright\arraybackslash}p{(\columnwidth - 6\tabcolsep) * \real{0.2500}}
  >{\raggedright\arraybackslash}p{(\columnwidth - 6\tabcolsep) * \real{0.2500}}
  >{\raggedright\arraybackslash}p{(\columnwidth - 6\tabcolsep) * \real{0.2500}}@{}}
\toprule\noalign{}
\begin{minipage}[b]{\linewidth}\raggedright
Objection
\end{minipage} & \begin{minipage}[b]{\linewidth}\raggedright
Why it misses
\end{minipage} & \begin{minipage}[b]{\linewidth}\raggedright
It would bite only if\ldots{}
\end{minipage} & \begin{minipage}[b]{\linewidth}\raggedright
Argued in
\end{minipage} \\
\midrule\noalign{}
\endhead
\bottomrule\noalign{}
\endlastfoot
\textbf{\(C\) is description-language dependent} (Kolmogorov) & we claim
only the \emph{sign} of \(C-1\), which is \textbf{provably invariant}
under faithful re-description (up to the \emph{review} band) & \ldots we
claimed \(C\)'s value --- which we never do & §3.4, §8.4 \\
\textbf{Kolmogorov complexity is uncomputable} & we never compute \(K\);
only a pre/post count under a fixed language & \ldots the criterion
required true \(K\) --- it does not & §3.3, §8.1 \\
\textbf{Reflexivity --- ``true by whose criterion?''} & compression is
independently measurable, so the rubric is checked against a quantity
outside itself and unseen cases & \ldots the criterion were
self-certifying --- it is checked externally & §8.1, §8.3 \\
\textbf{Single-curator labels} & validity rests on measurement, not
inter-rater agreement: label ⊥ computation (T1) and author ⊥ applicator
(human-blind) & \ldots the labels were the ground truth --- but the
independent count is & §8.0, §6 \\
\textbf{Framing is a free parameter} (the applicator can invent a
permissive frame) & the verdict is robust under a \emph{fixed} frame;
the freedom to \emph{choose} one is bounded by framing discipline ---
use the record's own description, forbid favorable reframing, require
multi-judge agreement --- and a disciplined re-audit shows every flip is
applicator permissiveness, not a corpus error & \ldots a permissive
frame survived the discipline --- but 0/11 of the swept cluster do &
§8.4 \\
\end{longtable}

Two honest residues remain. First, the curator's labels are shared
across records and judgment enters the counting protocol, so we claim
measurement validity on the corpus, \textbf{not} full independence. A
separate labeling team (true inter-rater reliability) is future work
(§10).

Second, and sharper: invariance is \textbf{stratified}, and at the
\(C\approx1\) boundary it is genuinely imperfect. At \(n=50\),
sign-invariance under admissible re-expression is 0.85 --- 0.92 for
strong compression but \textbf{0.78 for near-boundary \texttt{not\_ci}},
missing the 0.90 bar (§8.4). We do not paper over this. Two things make
it a bounded threat rather than a refutation. (i) The flips are
\textbf{definitional, not an irreducible principle}: tightening the
under-specified gate terms (chiefly ``boundary eliminated'' vs
``frontier merely moved'') raised boundary invariance from 0.57 to 0.78
--- defining the term collapses the flip it governs. (ii) The remainder
is \textbf{framing discipline}: when the applicator may re-choose the
frame, rejection fragility worsens (TNR 0.70 vs identification 1.00),
yet a disciplined re-audit --- use the record's existing description,
forbid favorable reframing, require independent judges to agree ---
resolves every flip to applicator permissiveness, not a corpus error.
The honest scope of the objectivity claim is therefore \textbf{strong
compression under a disciplined frame}, not unconditional
language-neutrality; the boundary is framing-discipline-bounded, and we
report the residual rather than claim it away.

\hypertarget{limitations-future-work}{%
\section{§10 --- Limitations \& Future
Work}\label{limitations-future-work}}

\hypertarget{this-paper-within-the-program}{%
\subsection{10.1 This paper within the
program}\label{this-paper-within-the-program}}

This paper is the first step of a larger program. The eventual goal is a
unified account of creativity --- and, we suspect, of the same pattern
in life and intelligence --- as \textbf{compression}: the resolution of
a standing conflict into a representation strictly shorter than holding
its terms apart. That synthesis is \textbf{not asserted here}; it has to
be \emph{earned}, by establishing several independently testable pieces
and only then combining them:

\begin{enumerate}
\def\labelenumi{\arabic{enumi}.}
\tightlist
\item
  \textbf{What a creative integration is} --- a decidable definition
  that separates it from its look-alikes.
\item
  \textbf{Whether nature's creations carry the same structure} ---
  whether the gates, unrewritten, separate creative integration from its
  look-alikes among natural processes (e.g.~the evolution of an organ),
  or whether the category turns out to be confined to deliberate
  reasoning.
\item
  \textbf{Whether integrations that encode inner or expressive
  experience compress in the same sense} --- and if so, whether the same
  gates suffice or a second reading is doing part of the work.
\item
  \textbf{Whether the method can \emph{generate} integrations, not only
  judge them.}
\item
  \textbf{Whether applying the method yields better outcomes on real
  tasks} --- a practical-utility question, on a different evidential
  axis from the measurement validity of this paper.
\item
  \textbf{Whether the same boundary doubles as a taxonomy of
  problem-solving methods} --- whether the look-alikes rejected here are
  not merely errors but the appropriate moves for different conflict
  structures (eliminate a cause, separate independent axes, sequence or
  standardize, package decisions), with creative integration as the
  limiting case that \emph{dissolves} the conflict rather than
  separating or managing it.
\end{enumerate}

These are research themes, not results in waiting, and we state them as
questions on purpose. Each is separable enough to be settled or refuted
on its own, and what any of them turns out to establish --- or whether
the eventual claim is narrower than the question --- is not something
this paper can promise. The integrative synthesis, if the pieces support
one, belongs to a later treatment built \emph{on} whatever they
establish rather than asserted ahead of them.

Against that map, \textbf{the present paper establishes piece (1)} ---
the definition, its gates and pitfall boundary, and the
\textbf{measurement validity of the judgment} --- on a single clean
evaluation. Its boundaries are therefore the pieces it does \emph{not}
yet cover:

\begin{itemize}
\tightlist
\item
  \textbf{Judgment, not generation.} We decide whether a resolution is a
  creative integration; we do not produce one. That is piece (4).
\item
  \textbf{Breadth is shown, not claimed.} The corpus spans many domains,
  but here they serve as breadth and calibration. Nature's creations and
  expressive cases (themes 2--3) need arguing in their own right, in
  their own fields, and this paper does not do that.
\item
  \textbf{Utility is a separate axis.} Measurement validity does not
  establish that \emph{using} the method helps in practice; that is
  piece (5).
\item
  \textbf{Curation independence.} The corpus is author-curated and
  judged primarily by a single curator. We mitigate this with
  measurement validity (label ⊥ computation, author ⊥ applicator) rather
  than human inter-rater agreement --- a deliberate choice (§8.0, §9)
  --- and ship the instrument for two further human-label checks:
  \textbf{(a)} test-retest / intra-rater reliability (same curator,
  verdict-blind, rubric-only), runnable now; and \textbf{(b)} a true
  inter-rater check by an independent rater, which remains open as no
  independent rater is yet secured. Full independence is future work.
\end{itemize}

\hypertarget{discussion-a-conjecture-is-genuine-creativity-identical-to-ci}{%
\subsection{10.2 Discussion --- a conjecture: is genuine creativity
identical to
CI?}\label{discussion-a-conjecture-is-genuine-creativity-identical-to-ci}}

The constitutive reading of §2 invites a stronger, \emph{eliminative}
conjecture, which we state plainly as a target for refutation:
\textbf{genuine creativity \emph{is} creative integration --- what has
no integrative core is novel, perhaps surprising, but not creative.}
Three observations make the bet serious. (i) Across Boden's types, the
cases people robustly call creative carry a \emph{resolved tension},
whereas merely additive combinations or routine exploration --- novel
but not felt as creative --- do not. (ii) Even the apparent hard case,
an elegant new proof of a \emph{known} theorem, is integration: the
original proof is long (\(L_{\text{pre}}\) large), the elegant one
collapses it in a single structural move --- a homomorphism, a duality,
a bijection --- with \(C \gg 1\) located in exactly that operation (the
algebra/geometry bridge of the complex plane, our calibration case, is
of this kind). (iii) Schmidhuber's compression-progress already implies
a resolved tension --- the observer's current model versus a shorter one
--- in every act we find interesting.

The conjecture is \textbf{not} vacuous, because the locus gate (§4, G4)
keeps it falsifiable: a description that is merely shorter by better
encoding (codification, §5) is \emph{not} CI; only compression
\emph{located in an explicit integrating operation} counts. So ``elegant
⇒ creative ⇒ CI'' holds for the illuminating cases and fails for mere
re-encoding --- a line one can check, not assert. We have found no
genuinely creative act that is mere novelty without integration, and we
\textbf{invite the counterexample}: the criterion can adjudicate any
candidate. The discipline the conjecture demands is to keep the conflict
\emph{content-level} and gate-checked; were we to let it become purely
epistemic --- any shortening read post hoc as resolving ``an
expectation'' --- the claim would be true by construction, and
worthless.

This also sharpens what CI offers \emph{generation} (piece 4): for an
open problem, where no solution is yet known, CI is not a solver but a
\textbf{search heuristic} --- diagnose the conflict's type and try the
corresponding integrating operation (a representation bridge for two
structures that will not talk; an invariant for local-versus-global; a
product for two independent structures). It narrows the search without
guaranteeing the operation exists --- and when none does, the
impossibility can itself be a higher-order integration (as in the
account of which equations are solvable by radicals).

A corollary sharpens the agent-neutrality of §3.2. Because the criterion
judges an act's \emph{structure}, not its author's intent or experience,
\textbf{machine creativity becomes decidable} --- not, as in
perception-based evaluation, a matter of whether observers attribute it
(Colton 2008). If a language model resolves a hard problem by an
integration that compresses a real conflict --- a refactor that
collapses two requirements that had been fighting, so a class of special
cases disappears (\(C>1\) at the locus) --- the act is a CI on exactly
the test applied to humans, and is creative; a one-line fix that merely
removes the immediate cause is competent but, by G1, not. The criterion
thus separates \emph{creative} machine work from merely \emph{correct}
machine work, decidably --- and it is here, on real tasks, that the
search heuristic above and the practical-utility piece meet: an agent
that diagnoses a conflict and applies the integrating operation is doing
creative work whose value is independently checkable.

\hypertarget{conclusion}{%
\section{§11 --- Conclusion}\label{conclusion}}

Creative integration can be defined, judged, and verified as a decidable
compression event. \emph{What it is:} under a fixed description
language, a resolution of a real conflict \(A\oplus B\) is a CI exactly
when the description strictly shrinks (\(C>1\)), with the saving located
in the conflict itself. \emph{How it is secured:} four binary gates
decide it, a taxonomy of pseudo-integration bounds it by naming the
look-alikes it rejects, and --- crucially --- we establish it by
\textbf{measurement, not consensus}: four falsifiable tests that show it
tracks the quantity it approximates on both axes, process and outcome.

\emph{The payoff:} a citable primitive. With \emph{what a creative
integration is} now decidable --- and, we conjecture, what separates
genuine creativity from mere novelty (§10.2) --- the program's next
questions --- nature's creations, expressive scope, generation, and
practical utility --- each become testable on their own terms rather
than matters of taste. Creative integration moves \textbf{from a
metaphor to a measurable object} --- the first move in bringing
creativity itself within the domain of science.

\hypertarget{references}{%
\section{References}\label{references}}

\small
\setlength{\parindent}{0pt}
\setlength{\parskip}{2.5pt plus 1pt}

Boden, M. A. 2004. \emph{The Creative Mind: Myths and Mechanisms}. 2nd
ed.~London: Routledge.

Colton, S. 2008. Creativity Versus the Perception of Creativity in
Computational Systems. In \emph{Creative Intelligent Systems: Papers
from the AAAI Spring Symposium}, Technical Report SS-08-03, 14--20.
Menlo Park, CA: AAAI Press.

Fauconnier, G., and Turner, M. 2002. \emph{The Way We Think: Conceptual
Blending and the Mind's Hidden Complexities}. New York: Basic Books.

Friedman, M. 1974. Explanation and Scientific Understanding.
\emph{Journal of Philosophy} 71(1):5--19.

Hutter, M. 2005. \emph{Universal Artificial Intelligence: Sequential
Decisions Based on Algorithmic Probability}. Berlin: Springer.

Jordanous, A. 2012. A Standardised Procedure for Evaluating Creative
Systems: Computational Creativity Evaluation Based on What it is to be
Creative. \emph{Cognitive Computation} 4(3):246--279.

Kitcher, P. 1981. Explanatory Unification. \emph{Philosophy of Science}
48(4):507--531.

Kitcher, P. 1989. Explanatory Unification and the Causal Structure of
the World. In Kitcher, P., and Salmon, W., eds., \emph{Scientific
Explanation}, 410--505. Minneapolis: University of Minnesota Press.

Koestler, A. 1964. \emph{The Act of Creation}. London: Hutchinson.

Kolmogorov, A. N. 1965. Three Approaches to the Quantitative Definition
of Information. \emph{Problems of Information Transmission} 1(1):1--7.

Rissanen, J. 1978. Modeling by Shortest Data Description.
\emph{Automatica} 14(5):465--471.

Ritchie, G. 2007. Some Empirical Criteria for Attributing Creativity to
a Computer Program. \emph{Minds and Machines} 17(1):67--99.

Schmidhuber, J. 2010. Formal Theory of Creativity, Fun, and Intrinsic
Motivation (1990--2010). \emph{IEEE Transactions on Autonomous Mental
Development} 2(3):230--247.

Wiggins, G. A. 2006. A Preliminary Framework for the Description,
Analysis and Comparison of Creative Systems. \emph{Knowledge-Based
Systems} 19(7):449--458.

\hypertarget{appendix-a-worked-examples}{%
\section{Appendix A --- Worked
Examples}\label{appendix-a-worked-examples}}

We do not release the full corpus (§7.4); instead we work the four-gate
criterion (§4) here on cases a reader can adjudicate unaided ---
consensus integrations (A.1), a pair that shows the verdict tracks the
\emph{framing} rather than the object (A.2), and rejected look-alikes
(A.3). Each follows the recording format of §3.3 (four-category count)
and §4.3 (gate trace); the running example, Maxwell, is in the main
text.

\hypertarget{a.1-consensus-integrations-positives}{%
\subsection{A.1 Consensus integrations
(positives)}\label{a.1-consensus-integrations-positives}}

Each follows three beats --- the \emph{conflict} (why \(A\) and \(B\)
genuinely compete), the \emph{gate-passing integration} (four-category
count, \(C>1\)), and the \emph{tell} (a prediction or new reach a
re-description could not buy).

\textbf{The periodic table --- element individuality \(\oplus\) periodic
regularity.} \emph{The conflict:} by the mid-19th century
\textasciitilde60 elements were known, each with its own reactivity
(individuality), yet ordering by atomic weight hinted at recurring
properties (regularity); partial rules --- Döbereiner's triads,
Newlands' octaves --- held locally but failed globally, so the two
readings competed. \emph{The integration \& gates:} two principles ---
order by atomic weight, fold on chemical period --- place every element
at a (row, column). The count runs \(O(N)\) → \(O(1)\) (pre: a
per-element property list plus the failed partial rules as exceptions;
post: 2 principles + a coordinate), so G3 passes; G1 passes (neither
individuality nor periodicity is cause-eliminable), G2 passes (the
failed triads/octaves show they competed), G4 passes (the ``individual
or periodic?'' boundary and those exceptions dissolve into one table).
\emph{The tell:} the table \emph{predicts} the undiscovered Ga/Ge/Sc
from its gaps, confirmed later to \textasciitilde2\% --- compression
buying reach a re-description never could. → \textbf{ci.}

\textbf{Special relativity --- space \(\oplus\) time.} \emph{The
conflict:} Galilean velocity-addition (space and time absolute) cannot
coexist with Maxwell's invariant \(c\); the ether program patched the
gap with a growing stack of assumptions, and the
moving-magnet/moving-conductor asymmetry resisted. \emph{The integration
\& gates:} one spacetime with the invariant \(c\), from two postulates
--- the count runs ≈7 ether/frame assumptions → 2 (G3). G1 passes (the
\(c\)-vs-Galilean conflict is real), G2 passes (absolute simultaneity
cannot coexist with invariant \(c\) --- not ``do both''), G4 passes (the
absolute-frame boundary and the induction asymmetry vanish; simultaneity
becomes frame-relative --- the saving sits at the conflict). \emph{The
tell:} the unification \emph{predicts} time dilation and \(E=mc^2\),
phenomena outside the original dispute --- the new-reach signature of
genuine compression. → \textbf{ci.}

\textbf{The limit (rigorous calculus) --- infinitesimal intuition
\(\oplus\) rigor.} \emph{The conflict:} infinitesimals were intuitive
but inconsistent (Berkeley's ``ghosts of departed quantities'');
rigorous exhaustion was sound but opaque and argued case by case ---
intuition and rigor pulled apart. \emph{The integration \& gates:} one
concept --- the \(\varepsilon\)--\(\delta\) limit, the infinite captured
by a finite procedure. The count runs \(O(N)\) per-case exhaustion
arguments → one definition \(O(1)\) (G3); G1 passes (Berkeley's critique
bit --- the conflict was real), G2 passes (one could not keep both
separately), G4 passes (the intuition/rigor split dissolves --- the
limit \emph{is} both). \emph{The tell:} the limit does not merely
re-describe the old proofs; it \emph{generates} them and the rest of
analysis from one definition --- reach again. → \textbf{ci.}

\hypertarget{a.2-the-verdict-tracks-the-framing-not-the-object}{%
\subsection{A.2 The verdict tracks the framing, not the
object}\label{a.2-the-verdict-tracks-the-framing-not-the-object}}

The locus gate (§4, G4) has a sharp consequence: the \emph{same} object
can be a CI or a mere look-alike depending on how the conflict is cut.
We record both cuts of two cases --- the \emph{mis-cut} (reads as a
look-alike), the \emph{right cut} (the genuine \(A\oplus B\) that
passes), and the \emph{tell} that shows which is real.

\textbf{The complex plane.} \emph{Mis-cut:} as \emph{real \(\oplus\)
imaginary} --- ``\(x^2+1=0\) has no real root, so adjoin \(i\)'' --- the
reduction merely supplies a missing prerequisite (algebraic closure); G1
fails, the pattern is cause-elimination (§5), \textbf{not\_ci}.
\emph{Right cut:} as \emph{algebra \(\oplus\) geometry} --- equations
and roots versus the plane and continuity, two domains long held apart
--- the complex plane and Riemann surfaces identify a number with a
point and an algebraic operation with a conformal map; all gates pass,
\textbf{ci}. \emph{The tell:} the right cut keeps paying out ---
function theory, Fourier analysis, and Hilbert space all flow from the
same analytic structure --- while the mis-cut buys nothing beyond
closing the gap.

\textbf{The real line.} \emph{Mis-cut:} as \emph{rational \(\oplus\)
irrational} --- ``the rationals have gaps, so add the irrationals'' ---
cause-elimination again, \textbf{not\_ci}. \emph{Right cut:} as
\emph{number \(\oplus\) magnitude} --- arithmetic (ratios) versus
geometric magnitude (length), split for two millennia after the
discovery of incommensurability and kept apart by Eudoxus' theory of
proportion --- Dedekind's and Cantor's continuum \(\mathbb{R}\) makes
every point of the line a number; all gates pass, \textbf{ci}. \emph{The
tell:} the right cut enables the arithmetisation of analysis (real
analysis, topology, and measure follow), whereas the mis-cut only plugs
holes.

In both, the object did not change; the \emph{cut} did. A framing that
locates the saving in patching one side reads as cause-elimination; one
that locates a genuine \(A\oplus B\) and a saving \emph{at that
boundary} --- confirmed by the reach it buys --- reveals the CI. This is
why the criterion is gate-checked at the locus, and why §10.2 insists
the conflict stay content-level rather than retrofitted.

\hypertarget{a.3-rejected-look-alikes-pseudo-integration}{%
\subsection{A.3 Rejected look-alikes
(pseudo-integration)}\label{a.3-rejected-look-alikes-pseudo-integration}}

The criterion earns its discrimination by what it \emph{refuses} (§5).
Three rejections, each failing a specific gate --- and each showing a
different way a non-integration can masquerade as one.

\textbf{Cause-elimination --- fails G1 (the ``conflict'' was a missing
prerequisite).} Reinforcement learning from human feedback can be told
as an integration: an agent \emph{must act} but has \emph{no reward
signal}, and the method reconciles them. But the two sides never
competed --- one was simply absent. Train a reward model (supply the
missing input) and the tension is gone. G1 asks exactly this:
\emph{would supplying a missing prerequisite make the conflict
disappear?} Here it would. The tell: cause-elimination \textbf{adds} a
component to unblock the problem, whereas a CI \textbf{removes} boundary
and exception cost --- Maxwell adds no missing ingredient; it sees that
\(E\) and \(B\) were one field.

\textbf{Organizational packaging --- fails G4 (the saving is in the
filing, not the conflict).} Consider a \emph{release-tiering policy}
that sorts AI model deployments into capability tiers, each gated by its
own safety requirements. Bundling many scattered deployment calls under
one framework --- one document replacing many ad-hoc judgments --- the
four-category count can even read \(>1\). But G4 asks \emph{where} the
reduction lands. Inspect any tier boundary and the release-versus-harm
trade-off re-appears in full --- you still decide, case by case, whether
a capability crosses the line. The boundaries did not dissolve; they
were relabelled and grouped. Packaging compresses the \emph{paperwork};
a CI compresses the \emph{conflict} (after Maxwell there is no
``electric or magnetic?'' decision left to make at all).

\textbf{Dissolved paradox --- fails G1/G2 (there was no real conflict to
begin with).} A candidate is dressed up as a striking paradox ---
``\(A\) and \(B\) are irreconcilable!'' --- then ``resolved.'' For
example, ``an assistant must be both \emph{helpful} and \emph{harmless}
--- a deep paradox!'' dissolves on inspection: the two largely act on
\emph{different} requests (help on benign ones, refuse harmful ones), so
``just do both'' handles the bulk (G2), and what remains is a dial to
set, not a boundary that compresses into a shorter description. In
general the sides either act on different referents (G2) or rest on an
unstated premise that, once named, removes the tension (G1); the
candidate is rewarded for \emph{sounding} paradoxical. This is the
sharpest structure-versus-rhetoric test (§5.2): a rubric that accepts it
is scoring presentation, not compression. A genuine CI survives the same
scrutiny --- space and time really cannot both be absolute under an
invariant \(c\) --- and only then is it compressed.

\end{document}